# Keypoint Density-based Region Proposal for Fine-Grained Object Detection and Classification using Regions with Convolutional Neural Network Features

**JT Turner**[1], **Kalyan Gupta**[1], **Brendan Morris**[2], & **David W. Aha**[3]

[1] Knexus Research Corporation; 174 Waterfront Street Suite 310;
National Harbor, MD 20745
[2] Department of Electrical and Computer Engineering;
University of Nevada, Las Vegas; Las Vegas, NV 89154
[3] Navy Center for Applied Research in Artificial Intelligence;
Naval Research Laboratory (Code 5514); Washington, DC 20375

## Abstract

Although recent advances in regional Convolutional Neural Networks (CNNs) enable them to outperform conventional techniques on standard object detection and classification tasks, their response time is still slow for real-time performance. To address this issue, we propose a method for region proposal as an alternative to selective search, which is used in current state-of-the art object detection algorithms. We evaluate our Keypoint Density-based Region Proposal (KDRP) approach and show that it speeds up detection and classification on fine-grained tasks by 100% versus the existing selective search region proposal technique without compromising classification accuracy. KDRP makes the application of CNNs to real-time detection and classification feasible.

## 1. Introduction

Applications of image processing algorithms to Navy missions such as those involving intelligence surveillance and reconnaissance (ISR), maritime security, and force protection (FP) require that they achieve high accuracy and respond in real time. Conventional approaches to image classification tasks includes the use of keypoint descriptors and local feature descriptors (Lowe, 2004), which are binned into histograms and compared to other keypoints to match similarly featurized objects. For instance, the work of Felzenszwalb et al. (2010) on deformable part models and detection of parts gave rise to specialized part models that operate by transfer of likely locations (Goering et al., 2014), which achieved high classification and detection accuracy and speed on the fine-grained Caltech UCSD bird dataset (Wah et al., 2011). Recently, Convolutional Neural Networks (CNNs), a deep learning approach, has emerged as a promising technique that dramatically outperforms conventional approaches on classification accuracy. Evolving from the early work of Lecun et al. (1998), who primarily focused on image classification, CNNs can now achieve state-of-the-art performance on object detection tasks (Girshick et al., 2014). Although highly accurate they are still not suitable for real-time detection and classification applications. For instance, with R-CNN and Fast R-CNN (Felzenszwalb et al., 2010) the object detection pipeline takes 2 seconds.

Our goal in this paper is twofold. First, we develop approaches to speed up R-CNNs so that they perform at sub-second levels, and second, we examine the effectiveness of our approach on fine-grained detection and classification tasks. Fine-grained classification tasks focus on visually similar but semantically different categories and present substantially larger interclass ambiguities, which complicate detection and classification problems. Therefore, we propose a new technique, called *Keypoint Density-based Region Proposal* (KDRP), to include as an integral element of the R-CNN detection and classification pipeline. The existing evaluations and applications of R-CNN have focused on standard coarse (rather than fine-grained) detection and classification tasks (e.g., PASCAL VOC 2007/2012). However, naval applications often require fine-grained detection and classification (Morris et al., 2012).

Therefore, we evaluate KDRP on representative fine-grained data sets, namely UEC-100 food and CUB-200 birds. We found that KDRP provides a 100% speedup over Fast R-CNN without compromising detection and classification accuracy.

We structure the remainder of this paper as follows: We provide background on the Fast R-CNN detection pipeline in Section 2. In Section 3, we describe our KDRP algorithm, which replaces step 2 in the Fast R-CNN detection classification pipeline. Section 4 describes our experiment, along with the time and accuracy results. We discuss these further in Section 5, and then conclude with future work plans.

## 2. R-CNN Detection Classification Pipeline

Here we describe the processing pipeline of Fast R-CNN (Girshick, 2015), which we modify to reduce image processing response time. It includes the following steps: (1) Region Proposal (via KDRP); (2) Feature Extraction (via a trained network); (3) Non-Maximum Suppression; and (4) Hypothesis Selection. *Our proposed KDRP algorithm* (Section 3) *replaces the state-of-the-art algorithm (selective search) in Step 1*. We describe all four steps in the following subsections.

### 2.1 Region Proposal using KDRP

The purpose of region proposal is to generate enough regions such that there is a high probability that one of the regions $r$ in the set of all regions proposed $R$ will be an accurate region that bounds the ground truth object. Assuming we are given an image $I$ with dimensions $w \times h$, a **region** is a subset $r \subseteq I$, with dimensions $w' \times h'$, where $0 < w' \leq w,$ and $0 < h' \leq h$.

The state-of-the-art approach for region proposal is selective search (Uijlings et al., 2013), which is an image segmentation method that uses multiple image scales and eight opponent color-spaces (consisting of one luminance channel, and two color spaces stimulated by opponent cones in the retina) to generate regions that may contain target objects (in a true object detection task there can be any number from 0 to $n$ objects detectable in the image). This is a time-intensive approach because it creates several thousand regions to increase the likelihood that at least one proposed region contains the ground truth object (to enable detection). As we explain in Section 3, because KDRP never infers bounding boxes and cannot modify the boundaries of a bounding box, the region that contains the object must be generated or detection will fail. The implementation of Fast R-CNN (Girschick, 2015) uses selective search for region proposal. In Section 3 we will describe this method and explain how it differs from KDRP.

### 2.2 Feature Extraction

After region proposal a trained CNN generates features using image convolution, activation functions, and pooling. For a model trained on $n$ classes, the output from the classification layer comprises $n + 1$ probabilities. For the bounding box coordinates of these classes, there are $4n + 4$ coordinates. The reason for adding one to the number of classes is that it allows a "background" class. Each of the region's classification is the softmax of the $n + 1$ probabilities, and each class corresponds to a 4-tuple of coordinates. The computationally expensive convolutional phase needs to only occur once; in this implementation the Regions of Interest (ROIs) that are detected are passed through the network at the same time as the image itself, and translated to the ROI pooling coordinates. After the ROI pooling layer has been computed, the only computations that need to be repeated multiple times are the multiplications between the fully connected layers, and dense matrix multiplications have been optimized to be extremely fast on GPUs. The network topology is shown in Figure 1.

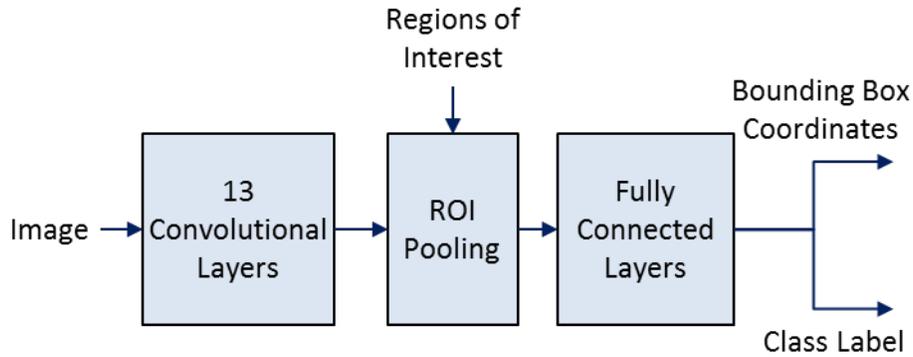

**Figure 1:** Feature extractor R-CNN's topology, modeled after the VGG16 architecture (Simonyan & Zisserman, 2014), with Regions of Interest (ROIs) and the final convolutional filters being given as input to the ROI pooling layer. The 13 convolutional layers, pooling layers, and Rectified Linear Unit (ReLU) layers are combined into one to increase the figure's readability.

*2.3 Non-Maximum Suppression*

Non-Maximum Suppression (NMS) (Girshick et al., 2014) is a technique for images containing an unknown number of objects. It combines likely detections of the same class together into one object if the detections overlap enough. NMS is very fast; it runs in $O(mcn^2)$ time, where $m$ is the number of images processed, $c$ is the number of classes, and $n$ is the number of regions. A description of NMS is given in (Girshick et al., 2014) and an example is shown in Figure 2.

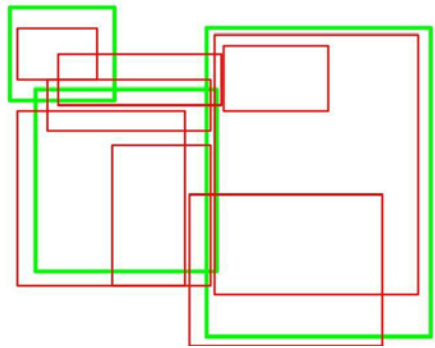

**Figure 2:** The three green regions are selected because they are the highest probability regions in the area that do not overlap at a fraction greater than $\alpha$ with a higher probability region. The red regions are suppressed because the regions they occupy were already occupied at a fraction greater than $\alpha$ by a higher probability region.

*2.4 Hypothesis Selection*

Hypothesis selection assigns class labels to regions of an image that have been featurized, or disregards the region. Like NMS, it is very fast with respect to region proposal or featurization. Hypothesis selection can be undertaken in one of two ways. In the first method, the algorithm is told the number of objects to detect in the image (e.g., Figure 3a shows a situation where it is told that there is one object in the image, when there are in fact 4). In the second (more realistic) method, the algorithm is not told how many objects (if any) are in the image, and it must determine which of its thousands of regions are valid detections using the confidence output from the R-CNN, as shown in Figure 3b.

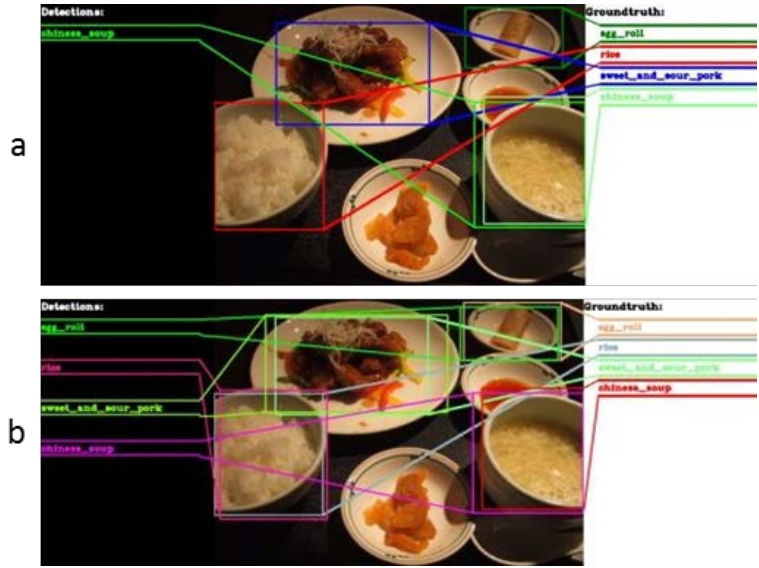

**Figure 3:** Comparison of two hypothesis selection methods, namely (a) top-k selection, or (b) selecting all detected objects greater than a known probability threshold. On the top, the system is told to select only one hypothesis (shown on the left) despite four ground truth objects being present. On the bottom, it is allowed to detect multiple objects, and all four ground truth objects are detected.

## 3. KDRP

Given the inefficiencies of selective search (Section 2.1), we propose an alternative for region proposal called KDRP (Algorithm 1). It generates $k$ arbitrary regions of an image in two steps:

---

KDRP(*image*, *regionsNeeded*) → *outputRegions*
1. *keypointCoordinates* ← SIFT-like feature generation(image)
2. *keypointMean* ←
      mean(numberOfKeypoints([256 uniform regions of *keypointCoordinates*]))
3. *keypointStdDev* ←
      stdDev(numberOfKeypoints([256 uniform regions of *keypointCoordinates*]))
4. *lengthOutput* ← 0
5. *outputRegions* ← [ ]
6. **while** (*lengthOutput* < *regionsNeeded*)
    6.1 *r* ← generateRandomRegion(*image*)
    6.2 *densityPercentile* ←
         percentile(zScore(numberOfKeypoints(*r*), *keypointMean, keypointStdValue*))
    6.3 **if** binomialTrialSuccess(*densityPercentile*)
        **then** *outputRegions* ← *outputRegions* + *r*
7. **return** *outputRegions*

**Algorithm 1**: Keypoint Density-based Region Proposal (KDRP) Algorithm

---

1. **Keypoint Generation *(Algorithm 1, Line 1)***: Line 1 computes SIFT features, which correspond to areas of a large change in the image's gradient. These areas are known to include distinguishing points of an image (Lowe, 2004). By focusing on regions with a high density of these descriptive keypoints, there is a greater likelihood that one region accurately bounds the target object. KDRP uses standard keypoint detection algorithms to find and plot points of interest on the image. The default SIFT-like features we use in step 1 are determined by cross validation using permutations of 8 keypoint methods on the UEC food dataset and the CUB bird dataset. The greatest performance boost is due to Shi Tomasi features, ORB features, and STAR features. This differs from the existing method of segmentation and nearest neighbor search of color segments, under the assumption that regions will contain strong corners and edges of the image, which is needed for recognition and detection.

2. *Region Hypothesis*: In line 2, a square window is slid over the given image using a uniform stride, and the density of key points from line 1 is sampled. After computing the mean and standard deviation of the region density, lines 6.1-6.3 stochastically generate a region (not of fixed width to height ratios), and examines how its density compares to the baseline just established. Regions in the $n^{th}$ density percentile are retained for detection following a binomial sampling with an $n\%$ chance of success. Sampling reduces bias of algorithms (Turner et al., 2014). This differs from traditional R-CNN with selective search in the time required to produce the equivalent number of regions.

A major advantage of KDRP versus selective search is that the exact number of regions to be used can be selected. The selective search implementation in MATLAB (as used in (Girshick, 2015; Girshick et al., 2014)) has a fast and slow mode, but the former cannot generate more regions than the slow mode. We conjecture that using relatively fast keypoint detection methods (Lowe, 2004) will be faster than the selective search method that generates and clusters opponent color spaces. An example of regions generated with KDRP is shown in Figure 4a, while a different image generated using selective search is shown in Figure 4b.

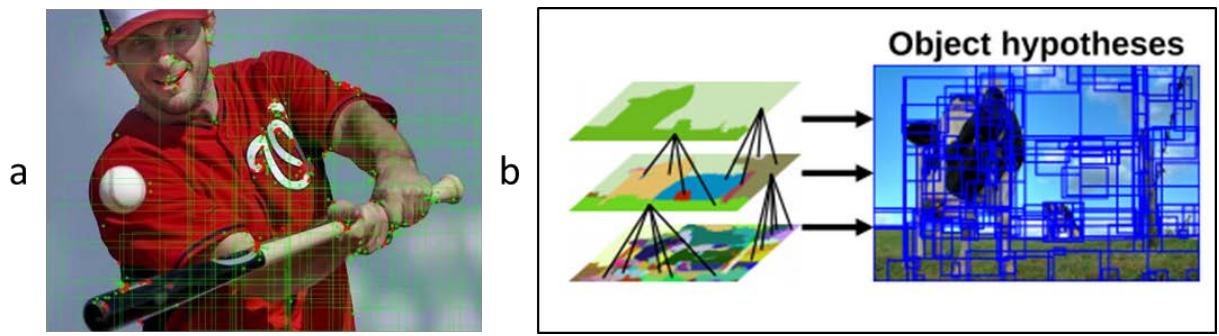

**Figure 4**: KDRP example of Max Scherzer batting. Red keypoints are ORB features, while green keypoints are STAR features. Only 5% of the regions are shown for increased visibility (Figure 4a), and color segmentation used to generate selective search regions is shown in Figure 4b.

## 4. Experiments

### 4.1 Objective

Our evaluation objective is to assess the performance advantages of KDRP versus selective search on fine-grained detection and classification tasks. Our metrics (see Section 4.2.3) are accuracy and response time.

### 4.2 Evaluation Method

#### 4.2.1 Datasets

We evaluate the two region proposal algorithms on the following two datasets (see Table 1). Since both datasets include bounding box coordinates, they are suitable for detection and classification evaluation. They offer unique challenges for detection above and beyond being fine-grained:

1. *UEC-100 food dataset*: This contains pictures of 100 categories of Japanese food, with bounding box annotations provided. UEC-100 has a variable number of ground truth objects per image, so we cannot hard code the algorithm to search for a pre-specified number of items. We discussed this issue in Section 2.4. Some example categories are Miso Soup, Chinese Soup, Soba Noodle, and Udon Noodle.

2. *CUB-200 bird dataset*. This contains pictures of 200 bird categories, with bounding box annotations provided. This presents a notoriously difficult classification task due to its many fine-grained categories. Example categories include American Crow, Yellow Bellied Warbler, Rock Warbler, and Baltimore Oriole.

**Table 1:** Dataset attributes for UEC-100 and CUB-200

| Dataset | Train/Test split | Number of Classes | Instances per Image |
|---|---|---|---|
| UEC-100 (Matsuda et al., 2012) | 10205/2966 | 100 | Variable in [0, n] |
| CUB-200 (Wah et al., 2011) | 5994/5794 | 200 | 1 |

### 4.2.2 Algorithms

We compared our implementations of the following two region proposal algorithms:

1. Selective Search Region Proposal (SSRP)
2. Keypoint Density-based Region Proposal (KDRP)

To test the two alternative region proposal approaches, we kept the detection-classification pipeline identical across the approaches except for line 1 (in Algorithm 1). However, the differences in the region proposal approaches do affect the downstream processes of feature extraction, NMS, and hypothesis selection.

For feature extraction, we separately trained and fine-tuned the pre-existing ImageNet model VGG16 (Simonyan & Zisserman, 2014), modifying only the output layers of the classification and bounding box regression steps. No other layers were modified (fully connected or convolutional). Fine-tuning was performed over the course of 5,000,000 iterations with a base learning rate of .01, decreasing by a factor of 10 every 500,000 iterations. The momentum term was set to .9, and weight decay was set to .0005.

During training, proposed regions were generated by the SSRP and KDRP algorithms, respectively. Using Uijlings et al.'s (2013) MATLAB code, selective search was applied to generate region proposals. The number of regions per image varied, as it depends on color features of the image, and averaged to around 2,000 per image. The number of regions generated by KDRP was set to 2,250; this was the maximum number of regions that could be used without the total time exceeding our predetermined threshold of 1 second per image. The study of the effect of using fewer regions per image warranted investigation, and we discuss this in Section 4.3.

Featurization has no tunable parameters, NMS was applied for high probability windows with an overlap exceeding 30% (as done by Girshick et al. (2014)), and the threshold for selecting a region for hypothesis selection when there is an unknown amount of objects in the ground truth was set to $p > .88$. This value was determined through cross validation only using the training set.

### 4.2.3 Metrics

We used the following two measures to compare the algorithms' performance:

1. *Time* ($t$): The response time taken in seconds to process a test image.
2. *Detection and classification accuracy* ($a$): We consider an object to be correctly detected if the intersection over union of the ground truth and bounding boxes is greater than 50%. Correctly identified ground truths are True Positives (TP), ground truths not covered by at least 50% of the detection box (or of the wrong class) are False Negatives (FN), and bounding boxes that do not correctly detect exactly one object are False Positives (FP). Average accuracy is defined as follows, for $x \in \{SSRP, KDRP\}$:

$$\mu a_x = \frac{TP}{TP + FP + FN}$$

### 4.2.4 Hypothesis Testing

We tested the following two null hypotheses using standard Student's $t$-tests.

1. KDRP's pipeline is faster than SSRP's pipeline: $H_0$: $\mu t_{KDRP} < \mu t_{SSRP}$, where $\mu t_{KDRP}$ is the mean image processing time for the KDRP pipeline and $\mu t_{SSRP}$ is the mean image processing time for selective search.
2. There is no difference between the detection accuracies of the KDRP and SSRP pipelines. $H_2$: $\mu a_{KDRP} = \mu a_{SSRP}$ where $\mu a_{KDRP}$ is the mean accuracy for the KDRP pipeline and $\mu a_{SSRP}$ is the mean accuracy for selective search.

### *4.3 Results*

The results of our evaluation are summarized in Table 2, which shows that KDRP's image processing response time is less than half of SSRP's for the two data sets (.96 seconds versus 1.96 seconds, and .968 versus 1.997 seconds). Therefore, we accept Hypothesis $H_0$.

Also, KDRP's detection accuracy does not differ significantly from SSRP's (68.03 versus 68.19, $p = .54$) for the UEC data set but is significantly higher for the CUB dataset (66.24 versus 65.18, $p = .0009$). Thus, Hypothesis $H_1$ is not supported.

We conclude that KDRP increases speed by roughly 100% versus SSRP without compromising detection accuracy. Furthermore, it reduces the overall processing time per image to under a second, which permits near real-time processing (a requirement for our target applications).

**Table 2:** Mean time and detection accuracy of the KDRP and SSRP pipelines

| Measures | Processing Time | | | | Detection Accuracy | | |
|---|---|---|---|---|---|---|---|
| Dataset | KDRP | SSRP | $n$ | $p$ | KDRP | SSRP | $p$ |
| UEC | .960 | 1.960 | 2966 | < .0001 | 68.03 | 68.19 | .54600 |
| CUB | .968 | 1.997 | 5794 | < .0001 | 66.24 | 65.18 | < .0001 |

## 5. Discussion

In Section 2.1 we argued that region proposal is the most time-intensive step in the detection and classification pipeline. Table 3 shows that it is as high as 63% for KDRP and 86% for SSRP. Consequently our efforts in reducing the overall computation time were well motivated.

**Table 3:** Region proposal computation time as a percentage of total processing time

| Dataset | KDRP | SSRP |
|---------|------|------|
| UEC | 63.0% | 85.0% |
| CUB | 62.5% | 86.1% |

Unlike SSRP, KDRP provides the ability to control the number of regions proposed. We reported the results in Table 2 with a setting of 2250 proposals. We explored the relationship between the number of proposed regions and accuracy. The results of our study are shown in Figure 5. There is steady but small decrease in accuracy until reducing from 2250 to 500 proposals, after which there is a steep decrease. The overall computational time at 500 proposals for one image is .361 seconds for CUB-200 (region proposal accounts for 43% of total time), and .372 seconds for UEC-100 (45%).

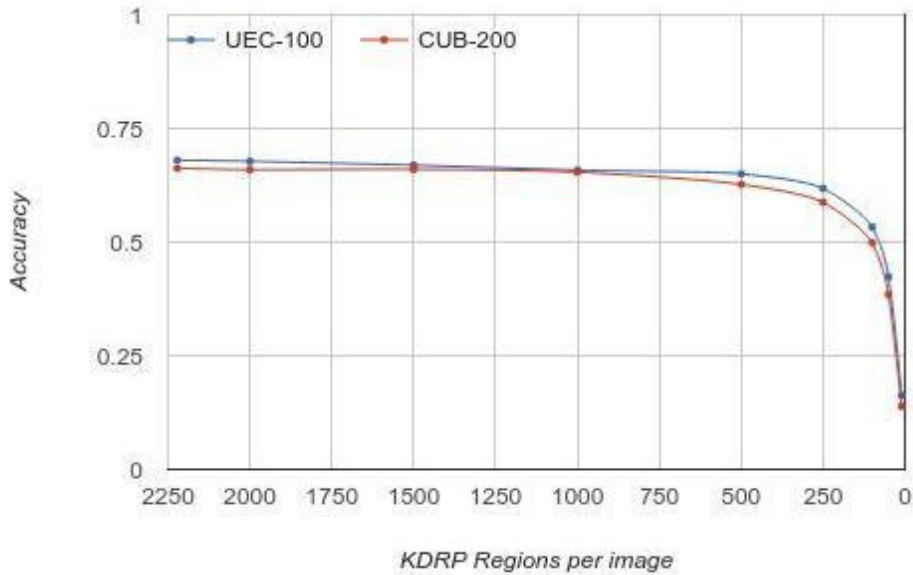

**Figure 5:** Effect of proposing fewer regions per image on detection accuracy for UEC-100 and CUB-200

## 6. Conclusion

Navy applications that require detection and classification of objects in images demand high accuracy and real-time response. However, while deep networks promise high levels of accuracy, they cannot yet respond in real-time. In this paper, we presented and evaluated a new approach called KDRP that modifies the current deep network approach to substantially reduce the response time without compromising accuracy. Furthermore, we evaluated our approach on fine-grained classification tasks that are relevant to naval decision making. Our goal was to develop a pipeline that places accurate bounding boxes around objects of interest in a dataset in under a second, so that it could be used in a low frame per second surveillance camera setting. We demonstrated that this goal is practically achievable with currently available hardware.

In our future work, we will investigate methods for improving KDRP by adding different region aspect ratios or scales upon acceptance. The nature of Fast R-CNN networks makes additional convolutions take marginal time, and increasing the number of regions proposed can only increase accuracy.

## Acknowledgements

Thanks to ONR for supporting this research.